\documentclass[10pt,twocolumn,letterpaper]{article}

\usepackage{cvpr}              % To produce the CAMERA-READY version
\usepackage[accsupp]{axessibility}  % Improves PDF readability for those with disabilities.
\usepackage[dvipsnames]{xcolor}

\usepackage{newfloat}
\usepackage{listings}
\usepackage{colortbl}
\usepackage{booktabs}
\usepackage{amsmath}
\usepackage{amssymb}
\usepackage{pifont}
\usepackage{multirow}
\usepackage{mathtools}
\usepackage{subcaption}

\newcommand\blfootnote[1]{%
\begingroup
\renewcommand\thefootnote{}\footnote{#1}%
\addtocounter{footnote}{-1}%
\endgroup
}

\definecolor{cvprblue}{rgb}{0.21,0.49,0.74}
\usepackage[pagebackref,breaklinks,colorlinks,citecolor=cvprblue]{hyperref}

\def\paperID{30} % *** Enter the Paper ID here
\def\confName{CVPR\xspace}
\def\confYear{2025\xspace}

\title{DuoSpaceNet: Leveraging Both Bird's-Eye-View and Perspective View Representations for 3D Object Detection}

\author{
    Zhe Huang$^{1,*}$ \quad 
    Yizhe Zhao$^{2,*}$ \quad 
    Hao Xiao$^{3}$ \quad 
    Chenyan Wu$^{4,\dagger}$ \quad 
    Lingting Ge$^{2}$\\[0.2cm]
    $^{1}$Carnegie Mellon University \quad 
    $^{2}$University of California, San Diego \quad \\
    $^{3}$University of Washington, Seattle \quad 
    $^{4}$The Pennsylvania State University \\[0.2cm]
    {\tt\small zhehuang@cmu.edu, yiz086@ucsd.edu, alexinuw@uw.edu, czw390@psu.edu, gelingting@gmail.com}
}

\begin{document}
\maketitle
\blfootnote{
    $^{*}$Equal contribution. $^{\dagger}$Corresponding author. This work was done at TuSimple (now CreateAI).
}

\begin{abstract}
Multi‐view camera‐only 3D object detection largely follows two primary paradigms: exploiting bird’s‐eye‐view (BEV) representations or focusing on perspective‐view (PV) features, each with distinct advantages. Although several recent approaches explore combining BEV and PV, many rely on partial fusion or maintain separate detection heads. In this paper, we propose DuoSpaceNet, a novel framework that fully unifies BEV and PV feature spaces within a single detection pipeline for comprehensive 3D perception. Our design includes a decoder to integrate BEV–PV features into unified detection queries, as well as a feature enhancement strategy that enriches different feature representations. In addition, DuoSpaceNet can be extended to handle multi‐frame inputs, enabling more robust temporal analysis. Extensive experiments on the nuScenes dataset show that DuoSpaceNet surpasses both BEV‐based baselines (e.g., BEVFormer) and PV‐based baselines (e.g., Sparse4D) in 3D object detection and BEV map segmentation, verifying the effectiveness of our proposed design.
\end{abstract}

\captionsetup{font=footnotesize}
\begin{figure}[!t]
\centering
\includegraphics[width=0.77\linewidth]{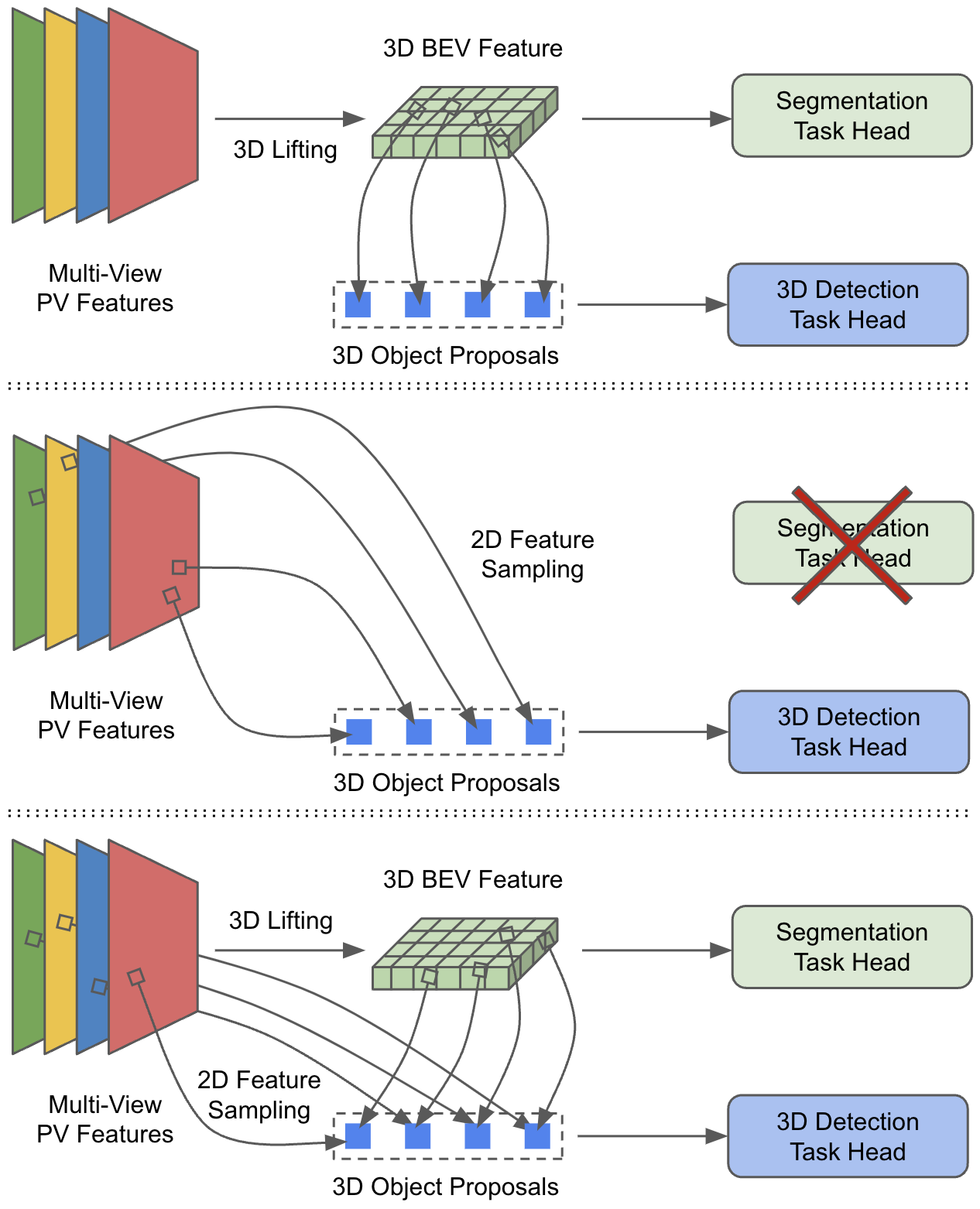}
\caption{Comparison of different image-only 3D perception frameworks. 
\textbf{(a) Top:} Bird's-eye-view-based (BEV-based) methods.
\textbf{(b) Middle:} Perspective-view-based (PV-based) 3D detection-only methods.
\textbf{(c) Bottom:} DuoSpaceNet (ours) where 3D detection benefits from both 3D BEV and 2D PV feature space.
}
\label{dense_vs_sparse}
\vspace{-0.3cm}
\end{figure}

\begin{figure*}[t]
    \centering  
    \includegraphics[width=0.75\linewidth]{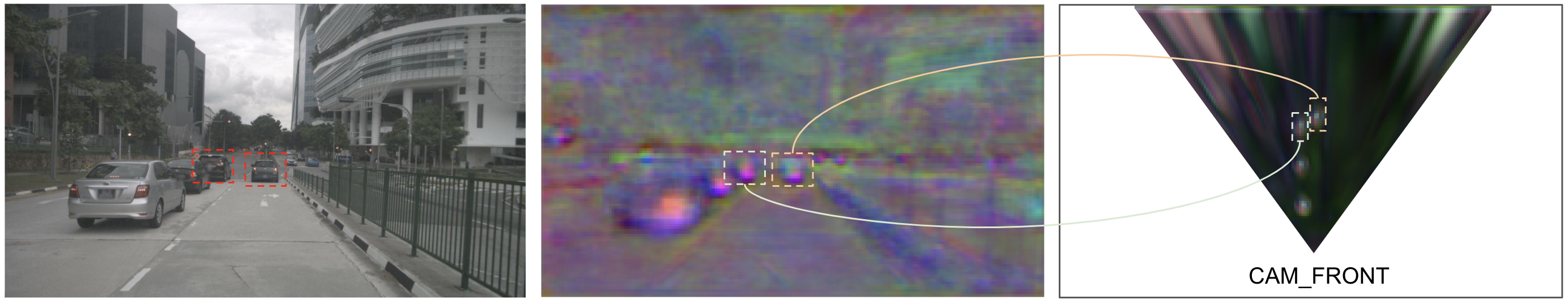}
    \vspace{-0.3cm}
    \caption{
    \textbf{Comparison of bird’s‐eye‐view (BEV) and perspective‐view (PV) features.}
    \textbf{(a) Left:} The original front-facing RGB image.
    \textbf{(b) Middle:} PV heatmap.
    \textbf{(c) Right:} BEV heatmap.
    Both heatmaps are produced by our final model.
    Different hues and brightness levels represent various response intensities with black indicating minimal or no response. 
    We highlight two car instances, each shown with its original image patch as well as corresponding PV and BEV heatmaps.
    The BEV heatmap in (c) makes it easier to interpret 3D positions---for instance, the relative positions of leading vehicles---without overlap issues.
    In contrast, the PV heat map in (b) preserves finer semantic details at higher resolution, which benefits attribute prediction.
    }
    \label{fig:raw_vs_pv_vs_bev}
% \vspace{-0.3cm} 
\end{figure*}

% \vspace{-0.5cm} 
\section{Introduction}
3D detection using multi‐view images represents an active area of research~\cite{wang2022detr3d, li2022bevformer, huang2022tig, xie2022m2bev, huang2021bevdet, huang2022bevdet4d, yang2023bevformerv2, liu2023sparsebev, lin2023sparse4dv2, qi2024ocbev, liu2022petrv2, lin2023sparse4dv3, wang2023streampetr, hu2023planningoriented}, driven by critical applications in autonomous driving~\cite{huang2022distributed} where it substantially improves downstream tasks such as object tracking~\cite{tang2018single, li2023end} and motion prediction~\cite{gan2023mgtr, zhang2023sapi,yu2021towards,yu2020data}. Although LiDAR‐based methods~\cite{yin2021center, chen2023voxelnext, ye2023lidarmultinet} often deliver superior performance for 3D detection, camera‐only solutions provide distinct advantages: they are typically more cost‐effective to deploy, perform reliably in adverse weather conditions such as rain or snow~\cite{yu2019gradual}, and offer higher resolution at greater distances with long focal length.

Most existing multi‐view camera‐only 3D object detection methods~\cite{wang2022detr3d, li2022bevformer, liu2022petrv2, lin2022sparse4d, han2023videobev} fall into two broad categories: \textit{bird’s‐eye‐view}‐based (BEV‐based) and \textit{perspective‐view}‐based (PV‐based), as illustrated in \cref{dense_vs_sparse}. BEV‐based methods~\cite{li2022bevformer, han2023videobev} generate a bird’s‐eye‐view feature map through a 2D‐to‐3D lifting process—often leveraging camera projection or unprojection—where each element of the BEV map corresponds to a specific 3D location. Conversely, PV‐based methods~\cite{wang2022detr3d, lin2022sparse4d} construct 3D detection heads atop 2D image features, typically relying on a limited set of sparse detection proposals. Visual cues are derived from the sparse feature locations where these proposals are anchored. In general, PV‐based methods prioritize object detection efficiency and computational speed.

While both BEV‐based and PV‐based methods can be effective, each exhibits notable limitations. In BEV‐based methods, multi‐view pose parameters guide the feature‐lifting process, enhancing 3D geometric awareness and ensuring robustness to object‐scale variations. They also avoid overlap artifacts arising from perspective projection. However, subtle visual cues can be lost due to coarse grid resolutions or interpolation. Moreover, the fixed‐size nature of the BEV feature map discards objects outside its predefined coverage, even if they are rapidly approaching the ego‐vehicle.
By contrast, PV‐based methods often maximize efficiency by sparsely attending to 2D PV features, allowing for theoretical detection at any distance. Yet, these methods typically rely on implicitly inferring 3D cues from 2D features, leading to weaker 3D perception. As shown in \cref{fig:raw_vs_pv_vs_bev}, retrieving spatial information from PV features is more difficult because of occlusion and perspective distortion, whereas BEV provides an unobstructed, rectified view. Still, BEV maps may omit fine‐grained details for small objects (e.g., pedestrians or cyclists) and are subject to camera ray‐like artifacts and limited coverage—issues that PV features do not face.

Some methods have attempted partial fusion of BEV and PV representations, yet they often depend on separate detection heads or derive visual cues predominantly from PV features. For example, BEVFormerv2~\cite{yang2023bevformer} combines BEV and PV features but maintains distinct detection heads, while the PETR series~\cite{liu2022petr, liu2023petrv2} focuses on integrating BEV positional encodings into PV features. These partial approaches underscore the need for a more holistic solution to fully harness the complementary strengths of BEV and PV.

We introduce \textit{DuoSpaceNet}, a holistic 3D perception framework that bridges the gap between bird’s‐eye‐view (BEV) and perspective‐view (PV) approaches by leveraging both representations for 3D object detection and map segmentation. The core novelty of DuoSpaceNet is its truly integrated dual‐space (also written as duo space) representation, where each object query contains both BEV and PV embeddings. Unlike systems that merely combine features from separate heads, our method preserves the distinctiveness of each representation while merging them in a single, unified decoder for robust 3D perception.

Our architecture retains BEV and PV features and processes them with the \textit{Duo Space Decoder}—a modified version of the transformer decoder from Deformable DETR. Each object query includes dual content embeddings from both spaces, along with a shared pose embedding representing the object’s real‐world 3D position. Within the decoder, space‐specific cross‐attention layers refine dual‐space object queries, enabling synergy between geometric and semantic cues. We further propose a feature divergence enhancement that maximizes the distinctiveness of BEV and PV features, leading to more discriminative representations.

Beyond singe-frame detection, DuoSpaceNet supports a temporal extension that elegantly infuses multi‐frame inputs into temporal queries, demonstrating our framework’s adaptability. For map segmentation, we append a segmentation head after BEV feature generation, with separate channels for each map category. By fusing both geometric and semantic insights in a single pipeline, DuoSpaceNet offers a comprehensive solution for multi‐task 3D perception, setting it apart from existing methods that rely on partial fusion or separate detection heads.

Our main contributions are as follows:
\begin{itemize}
\item \textit{Integrated Duo Space Framework}: We present \textit{DuoSpaceNet}, which unifies BEV and PV representations within a single detection pipeline. This design features a \textit{duo space decoder} for comprehensive BEV–PV fusion and a \textit{divergence feature enhancement} technique that preserves the strengths of each view while enriching the combined representation.

\item \textit{Temporal Extension}: Our method also extends to temporal dimension with motion-compensated temporal queries, assisting 3D detection with multi‐frame context.

\item \textit{Empirical Validation on nuScenes}: Through extensive experiments on the nuScenes~\cite{caesar2020nuscenes} dataset, DuoSpaceNet outperforms both PV‐only and BEV‐only baselines, as well as other strong methods, in 3D detection and segmentation. Ablation studies further confirm the effectiveness of each proposed component.
\end{itemize}

\section{Related Work}

\subsection{BEV-Based Multi-View 3D Perception}
Tackling multi-view 3D object detection using bird's-eye-view (BEV) representations has become popular in autonomous driving. Methods like LSS~\cite{philion2020lift}, BEVDet~\cite{huang2021bevdet,huang2022bevdet4d}, and BEVDepth~\cite{li2023bevdepth} project 2D image features into BEV feature maps using dense depth predictions. M${}^2$BEV~\cite{xie2022m2bev} and SimpleBEV~\cite{harley2023simplebev} enhance efficiency by assuming uniform depth during back-projection. Notably, BEV methods such as BEVFormer series~\cite{li2022bevformer, yang2023bevformerv2, qi2024ocbev} avoid dense depth estimation by modeling BEV features with dataset-specific queries via deformable attention~\cite{zhu2020deformable}. BEVFormer v2~\cite{yang2023bevformerv2} incorporates an auxiliary 3D detection head, while OCBEV~\cite{qi2024ocbev} focuses on instance-level temporal fusion. Recent methods~\cite{park2022solofusion, park2022time, han2023videobev, zong2023hop} primarily improve temporal modeling rather than fundamental BEV representations.

BEV-based methods also address map segmentation tasks. VED~\cite{lu2019ved} uses a variational encoder-decoder to generate BEV features, and VPN~\cite{pan2020vpn} employs an MLP-based transformation. Transformer-based approaches~\cite{zhou2022cvt, peng2023bevsegformer} utilize cross-attention mechanisms to produce BEV representations. Recent geometric reasoning methods such as LSS~\cite{philion2020lift} explicitly estimate depth distributions, while methods like M${}^2$BEV and BEVFormer leverage camera parameters and multi-task learning for enhanced detection and segmentation performance.

\subsection{PV-Based Multi-View 3D Perception}

Starting with DETR3D~\cite{wang2022detr3d}, the landscape of multi-view perspective-view-based (PV-based) 3D object detection leans towards sparse query refinement with set-to-set matching loss. Following studies like PETR~\cite{liu2022petr,liu2022petrv2} and CAPE~\cite{xiong2023CAPE} improve transformer-based detection decoders with 3D position-aware image features. Recently, Sparse4D~\cite{lin2022sparse4d} further extends this track with the introduction of 4D anchors, allowing intuitive ego-motion and object motion compensation. We also find it useful and derive our query design from it. 
Similar to the BEV-based 3D detection track, current state-of-the-art PV-based methods, such as~\cite{lin2023sparse4dv2, wang2023streampetr, liu2023sparsebev}, also focus on improving temporal modeling techniques by temporal query propagation.

\subsection{Unifying BEV and PV Detection}

While several approaches have explored the fusion of Bird's-Eye-View (BEV) and Perspective-View (PV) features for 3D detection, our proposed method stands apart by employing a fully shared detection head. BEVFormerv2 \cite{yang2023bevformer} pioneered the combination of BEV and PV queries to enhance BEV detection performance, but maintains separate detection heads for BEV and PV tasks. Furthermore, their detection queries contain either PV or BEV latent features exclusively, rather than integrating both.
The PETR series \cite{liu2022petr, liu2023petrv2} and StreamPETR \cite{wang2023exploring} represent an alternative approach utilizing 3D queries and 2D PV features to detect 3D objects. However, these methods differ fundamentally from our work as their 3D queries contain only position embeddings, with all visual information derived solely from 3D PV image features. In contrast, our 2D-3D fused queries explicitly generate and incorporate BEV features, obtaining more comprehensive cues from both feature representations.
MVF \cite{zhou2020end} presents perhaps the closest parallel to our approach, proposing a LiDAR-based 3D detection pipeline that leverages features from both 3D BEV space and 2D range image space. However, their method does not naturally extend to query-based camera-only detection systems like ours.

\begin{figure*}[!h]
	\centering  
     \includegraphics[width=0.95\linewidth]{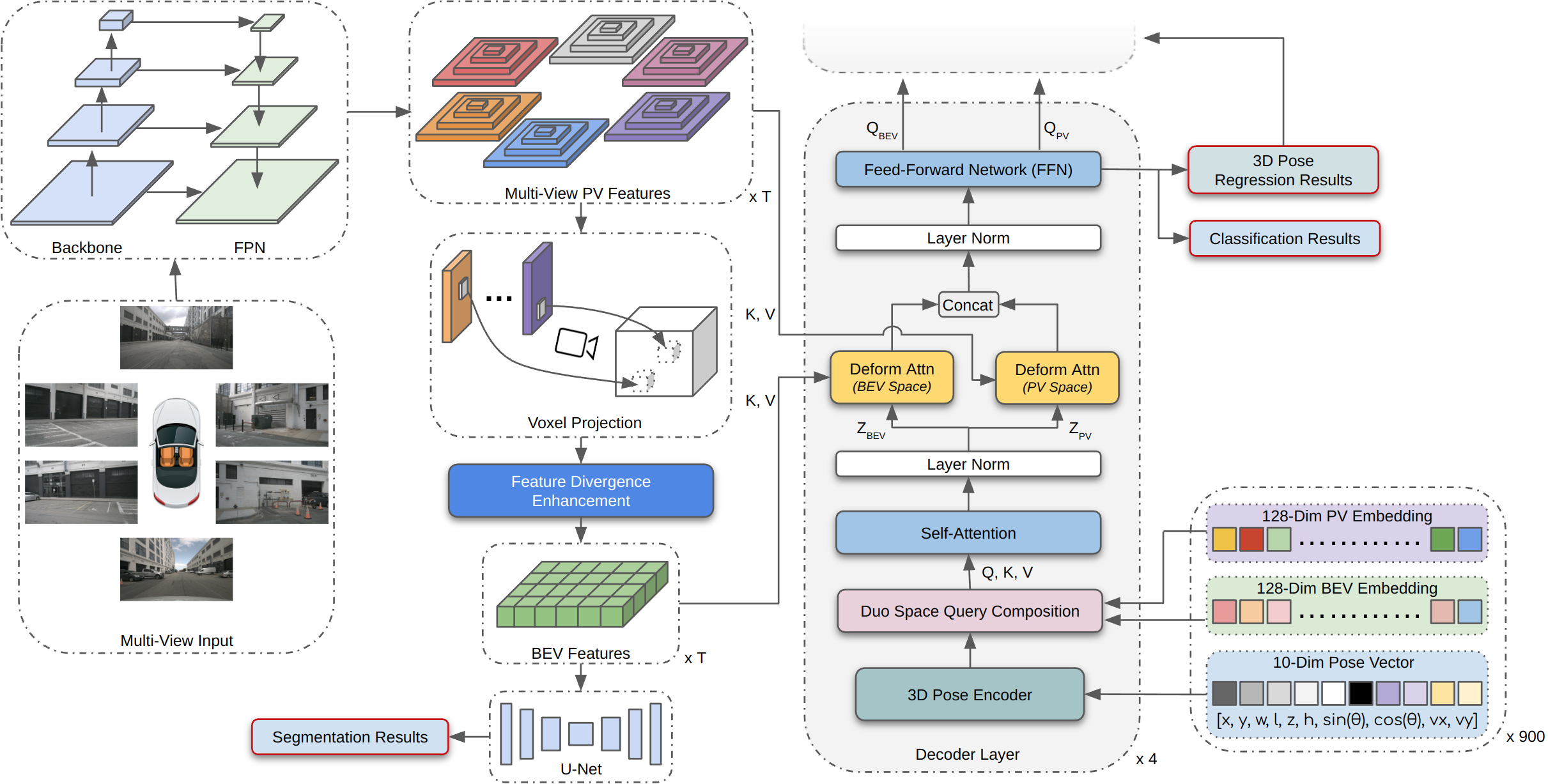}
 \caption{Overall architecture of the proposed DuoSpaceNet. Multi-view 2D perspective view (PV) features are extracted by the backbone and the feature pyramid network (FPN)~\cite{lin2017feature}. 
 Our 2D to 3D BEV lifting strategy consists of a parameter-free voxel projection following~\cite{harley2023simplebev} and a divergence enhancement process to make resulting BEV features more distinctive w.r.t. PV features.
 In our duo space framework, multi-view PV features and BEV features are identified as equally important and are fed into the decoder together.
 Each decoder layer has one self-attention layer and two deformable cross-attention layers~\cite{zhu2020deformable}.  The self-attention layer acts on both BEV and PV spaces, whereas each cross-attention layer only attends to either BEV features or PV features. This space-specific cross-attention helps preserve the uniqueness of different feature spaces throughout multi-layer refinement process. 
 Details about the Duo Space Query Composition can be found in \cref{Eq:z_bev,Eq:z_pv,Eq:qkv}.
 Dense map segmentation can be jointly carried out via a separate segmentation head.}
	\label{fig:framework}
\vspace{-0.5cm} 
\end{figure*}

\section{Method}
As shown in \cref{fig:framework}, in our framework, PV and BEV feature maps are firstly generated from multi-view images and are then fed into the duo space decoder for object detection. The object queries equipped with duo space embedding are updated layer-by-layer in the decoder and the outputs from the last layer are used for predicting object labels and 3D poses. In addition, a dense segmentation head is built upon the BEV feature map for map segmentation.

\subsection{Duo Space Features}
\paragraph{Feature extraction.} 
Multi-view images are first processed by an image encoder, which includes a backbone network (e.g., ResNet~\cite{he2016resnet}) and a neck (e.g., FPN~\cite{lin2017feature}), to generate multi-scale PV features $\{F_{PV}^j \in \mathbb{R}^{N \times C_j \times H_j \times W_j}, j=1,2,...M\}$, where $N, M$ are the number of cameras and different scales, $H_j, W_j, C_j$ denote the feature map height, width and channel number of the $j$-th scale. Multi-view multi-scale PV features are lifted from 2D to 3D via a simple parameter-free back-projection module following previous work~\cite{harley2023simplebev}. A 3D volume of coordinates with size $X \times Y \times Z$ is first generated and projected onto multiple images. PV features sampled around the projected positions are then aggregated by bilinear interpolation, resulting in a voxel feature map $F_{voxel} \in \mathbb{R}^{C \times X \times Y \times Z}$, where $X, Y, Z, C$ represent the size of the voxel space and channels of the voxel feature. Eventually, the $Z$ dimension is reduced to yield a BEV feature map $F_{BEV} \in \mathbb{R}^{C \times X \times Y}$.

In multi-frame settings, historical images are processed by the same procedure sequentially, generating PV and BEV feature maps for different frames. Both feature maps within a fixed temporal length are stored for future use. 

\vspace{-0.3cm}
\paragraph{Feature divergence enhancement.} 
Our model benefits from the contrastiveness between the two feature representations. 
Since our lifting method is parameter-free, its functionality can be viewed as rearranging PV features given priors (e.g., camera poses) on the 3D geometry of a scene. Therefore, it has minimal effects on diverging the feature contents. To increase the heterogeneity of BEV features w.r.t. PV features, we propose a simple yet effective divergence enhancement stage.

It consists of three 3D convolution layers (Conv3Ds) and three 2D convolution layers (Conv2Ds). 
First, we apply Conv3Ds on $F_{voxel}$ to improve 3D geometry awareness in a learning-based fashion. 
After $F_{voxel}$ is flattened along its $Z$ dimension, Conv2Ds are applied for further BEV-level refinement, yielding final $F_{BEV}$. 

\subsection{Duo Space Decoder} \label{sec_object_query}

\paragraph{Duo space queries.}
Suppose we have $k$ object queries, $\{Q^i\}_{i=1}^{k}$. Each consists of a pose embedding, $Q^i_{Pose}$, and duo space content embedding for both BEV and PV spaces, $Q^i_{BEV}$ and $Q^i_{PV}$, respectively. 

Adapted from~\cite{lin2022sparse4d}, each $Q^i_{Pose}$ is encoded from a 3D pose vector $\mathcal{P}_i$,  which contains attributes with physical meanings, including $x, y, z$ in the vehicle coordinate system, the width, length, height, orientation and the velocity of the object the query is associated with. While $Q^i_{BEV}$ and $Q^i_{PV}$ contain high-level content features in BEV space and PV space respectively. In each layer of the duo space decoder, first, a pose encoder consisting of several FC layers is used to encode $\{\mathcal{P}_i\}_{i=1}^{k}$ into high dimensional latent representations, dubbed  $\{\texttt{Enc}\left(\mathcal{P}_i\right)\}_{i=1}^{k}$, which will serve as learnable positional encodings in the subsequent attention layers. 

To unify the 3D pose of each object query across BEV and PV spaces, we generate a shared pose embedding,
\begin{align}
    Q^i_{Pose} = \xi\left(\texttt{Enc}(\mathcal{P}_i)\right), i \in \{1, 2, ..., k\},
\end{align}
where $\xi(\cdot)$ denotes a linear transformation to make the dimension of $\texttt{Enc}\left(\mathcal{P}_i\right)$ the same as $Q^i_{BEV}$ and $Q^i_{PV}$. The final duo space queries in BEV space and PV space can be derived by simply adding the corresponding content embedding with the shared pose embedding together by
\begin{align}
    \mathbf{z}_{BEV} &= \left\{Q^i_{BEV} + Q^i_{Pose}\right\}_{i=1}^k, \label{Eq:z_bev}\\
    \mathbf{z}_{PV} &= \left\{Q^i_{PV} + Q^i_{Pose}\right\}_{i=1}^k. \label{Eq:z_pv}
\end{align}
The self-attention layer thus can be represented as

\begin{gather}
    Q = K = V = \mathbf{z}_{BEV} \oplus \mathbf{z}_{PV}, \label{Eq:qkv} \\
    \texttt{MHSA}(Q, K, V) = \texttt{Softmax}\left(\frac{QK^T}{\sqrt{\texttt{dim}(K)}}\right)V, 
\end{gather}
where $\oplus$ denotes a concatenation operator along the channel dimension and $\texttt{MHSA}(...)$ stands for multi-head self-attention described in~\cite{vaswani2017attention}. 

\vspace{-0.3cm}
\paragraph{Space-specific cross-attention.} For multi-head cross-attention layers $\texttt{MHCA}_{BEV}(...)$ and $\texttt{MHCA}_{PV}(...)$, each of them will only act on their corresponding features using corresponding inputs. 

In BEV space, it can be represented as
\begin{align}
    \hat{\mathbf{p}}_{BEV} &= \left\{\mathcal{P}_i\vert_{x,y}\right\}_{i=1}^k, \label{Eq:p_bev}\\
    \texttt{MHCA}_{BEV}(...) &= \texttt{MSDA}\left(\mathbf{z}_{BEV}, \hat{\mathbf{p}}_{BEV}, F_{BEV}\right), \label{Eq:ca_bev}
\end{align}
where $\hat{\mathbf{p}}_{BEV}$ denotes the normalized coordinates of 3D reference points (only using their $X$ and $Y$ components here). $\texttt{MSDA}(...)$ is the Multi-Scale Deformable Attention Module (MSDeformAttn) described in~\cite{zhu2020deformable}. Similarly, we have cross-attention in PV space as
\begin{align}
    \hat{\mathbf{p}}_{PV} &= \left\{\texttt{Proj}\left(\mathcal{P}_i\vert_{x,y, z}, \{ \boldsymbol{K}_n, \boldsymbol{T}_n \}_{n=1}^{N}\right)\right\}_{i=1}^k, \label{Eq:p_pv}\\
    \texttt{MHCA}&_{PV}(...) = \texttt{MSDA}\left(\mathbf{z}_{PV}, \hat{\mathbf{p}}_{PV}, \{F_{PV}^j\}_{j=1}^M\right), \label{Eq:ca_pv}
\end{align}
where $\texttt{Proj}(...)$ refers to the projection of 3D LiDAR coordinates into 2D image frames using camera matrices $\{ \boldsymbol K_{n} \}_{n=1}^{N} \subset \mathbb{R}^{3 \times 3}$  and $\{ \boldsymbol T_{n} \}_{n=1}^{N} \subset \mathbb{R}^{4 \times 4}$. Since this attention happens in PV space, multi-scale PV features $\{F_{PV}^j\}_{j=1}^M$ are used. Following feature extraction and refinement through multi-head space-specific cross-attention layers, the outputs of $\texttt{MHCA}_{BEV}$ and $\texttt{MHCA}_{PV}$ are concatenated as refined object queries, which are then fed into a 2-layer feed forward network (FFN). Finally, the FFN outputs are used for object category prediction and are also decoded into a 10-dim 3D pose vector as our detection regression results.
The refined poses then serve as inputs for subsequent decoder layers. 

\begin{figure}
	\centering  
     \includegraphics[width=0.98\linewidth]{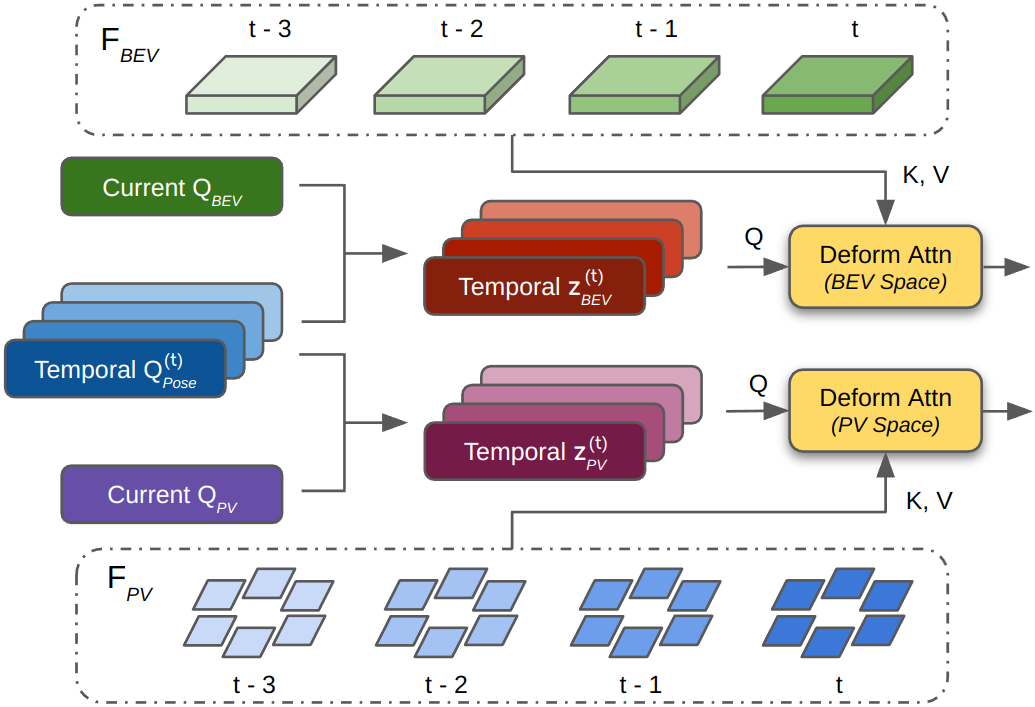}
 \caption{Diagram of the proposed duo space temporal modeling with 4 frames. Temporal pose embeddings $Q^{(t)}_{Pose}$ are generated by warping pose vectors at current timestamp through motion compensation. Subsequently, temporal duo space queries $\mathbf{z}^{(t)}_{BEV}$ and $\mathbf{z}^{(t)}_{PV}$ are assembled by broadcasting current content embeddings over the time dimension and then combining them with the temporal pose embeddings. We then conduct space-specific cross-attention using recent BEV and PV feature maps, both of which are maintained by their respective memory queue. Note that temporal queries from each timestamp only interact with feature maps corresponding to that timestamp. The resulting temporal queries are aggregated via a MLP in a recurrent fashion.}
	\label{fig:temporal}
\vspace{-0.5cm} 
\end{figure}

\subsection{Duo Space Temporal Modeling}
 
 BEV-based 3D detection methods~\cite{huang2022bevdet4d, li2022bevformer} typically utilize temporal inputs by stacking temporal BEV feature maps. Offsets are determined either with motion compensation or in a learnable manner (e.g., deformable attention) or both combined. Meanwhile, PV-based methods~\cite{lin2022sparse4d, liu2022petrv2} generally infuse temporal information into object queries. Therefore, the difference between BEV-based and PV-based temporal methods brings challenges to temporal design in our duo space paradigm. In this section, we present a unified temporal solution for both spaces via temporal duo space queries, illustrated in \cref{fig:temporal}. Concretely, for a fixed temporal length $l$, each object is represented by $l$ temporal duo space queries, each comprising a temporal pose of the underlying object and a shared content embedding. Temporal poses are deduced by applying both ego- and object-motion compensation on the object's current pose vector. 

 The queries are fed into the space-specific cross-attention layer corresponding to their space. Each query only attends to features from a specific timestamp associated with its temporal pose. Subsequently, results produced by $l$ temporal queries are recurrently aggregated via multi-layer perceptron (MLP) into a single refined prediction per object. Our solution elegantly operates symmetrically across BEV and PV spaces. Please refer to the supplementary materials for more details.

\subsection{Multi-Task Learning}
Similar to BEV-based methods, our model is capable of joint optimization of detection and segmentation. To perform dense segmentation, we simply add a segmentation branch consisting of a U-Net~\cite{ronneberger2015unet} like structure for feature enhancement and two parallel convolution-based segmentation heads for final predictions. It takes the BEV feature map $F_{BEV}$ as input, and outputs two segmentation masks of the same resolution. To supervise the map segmentation branch, we use a weighted sum of focal loss~\cite{lin2017focal} and dice loss~\cite{milletari2016vnet} during training.

\captionsetup{font=small}

\begin{table*}[!h]
\centering
% \caption{Comparison on 3D detection results on nuScenes \texttt{val} set. All methods are trained for 24 epochs using ResNet-101-DCN~\cite{zhu2018deformable} and benefit from perspective view pre-training. 
% % $\dagger$ indicates equivalent number of training epochs due to data augmentation. 
% Test time augmentation is not used for all experiments.
% }
% \label{tab:main_val_set}
\tiny
\resizebox{0.8\textwidth}{!}{
\setlength{\tabcolsep}{3.5pt}
\begin{tabular}{l|c|c|c|c|c|c@{\hspace{1.0\tabcolsep}}c@{\hspace{1.0\tabcolsep}}c@{\hspace{1.0\tabcolsep}}c@{\hspace{1.0\tabcolsep}}c@{\hspace{1.0\tabcolsep}}} 

\toprule
\textbf{Method} & \textbf{Image Size}  & \textbf{Frames} & \textbf{mAP}$\uparrow$  &\textbf{NDS}$\uparrow$  & \textbf{mATE}$\downarrow$ & \textbf{mASE}$\downarrow$   &\textbf{mAOE}$\downarrow$   &\textbf{mAVE}$\downarrow$   &\textbf{mAAE}$\downarrow$ \\
\midrule
% ResNet101
DETR3D \cite{wang2022detr3d} & 1600 $\times$ 900  & 1                  & 0.349 & 0.434 & 0.716 & 0.268 & 0.379 & 0.842 & 0.200 \\ 
% [sp4d erreta] BEVDet \cite{huang2021bevdet}   & 1600 $\times$ 900  & 1                   & 0.357 & 0.421 & 0.710 & 0.270 & 0.490 & 0.885 & 0.224 \\ % 
% PETR \cite{liu2022petr}   & 1408 $\times$ 512 & 1                                & 0.366 & 0.441 & 0.717 &   0.267 & 0.412 & 0.834 & 0.190\\ % PETR Table 1
PETR \cite{liu2022petr} & 1600 $\times$ 900 & 1                                & 0.370 & 0.442 & 0.711 &  0.267 & 0.383 & 0.865 & 0.201\\ % PETR Table 1
BEVFormer-S \cite{li2022bevformer}  & 1600 $\times$ 900  & 1                   & 0.375 & 0.448 & 0.725 & 0.272 & 0.391 & 0.802 & 0.200\\ % BEVFormer Table 4.
% BEVFormer v2 \cite{yang2023bevformerv2}  & 48 & 1600 $\times$ 900  & 1                   & 0.374 & 0.451 & 0.730 & 0.270 & 0.379 & \textbf{0.773} & 0.205\\ % BEVFormer Table 4.
Sparse4D \cite{lin2022sparse4d}  & 1600 $\times$ 640 & 1                 &   {0.382} & 0.451 & 0.710 & 0.279 & 0.411 & 0.806 &   0.196\\ % Sparse4d Table 2
% Focal-PETR \cite{wang2022focal}  & 1408 $\times$ 512 & 1                                & 0.390 & 0.461 & \textbf{0.678} & 0.263 & 0.395 & 0.804 & 0.202\\ % PETR Table 1
SimMOD \cite{zhang2023simple}  & 1600 $\times$ 900  & 1                   & 0.366 &   {0.455} &   {0.698} & 0.264 & 0.340 &   0.784 & 0.197\\ % SimMOD Table 1.
% CAPE \cite{xiong2023CAPE} & CVPR2023 & 1600 $\times$ 900 & 1 & 0.388 & 0.463 & - & - & - & - & - \\ % CAPE Table 3
\rowcolor[gray]{.9} 
DuoSpaceNet (Ours) & 1600 $\times$ 640 & 1                                 &\textbf{0.399}  &\textbf{0.462}  &0.683  &0.279  &  {0.376}  &0.829  &0.205\\
\midrule

% PolarDETR \cite{chen2022polar} & arXiv2022  & 1600 $\times$ 900  & 2                 & 0.383 & 0.488 & 0.707 & 0.269 & 0.344 & 0.518 & 0.196 \\ % PolarDETR Table 1
UVTR \cite{li2022unifying} & 1600 $\times$ 900   & 6                      & 0.379 & 0.483 & 0.731 & 0.267 & 0.350 & 0.510 & 0.200\\
% BEVDepth \cite{li2023bevdepth} & 90${}^\dagger$ & 1408 $\times$ 512 & 2                   & 0.418 & 0.538 & 0.565 & 0.266 & 0.358 & 0.331 &   {0.190} \\ % BEVDepth Table 7
BEVFormer \cite{li2022bevformer} & 1600 $\times$ 900  & 4                   & 0.416 & 0.517 & 0.673 & 0.274 & 0.372 & 0.394 & 0.198\\ % BEVFormer Table 4.
PETRv2 \cite{liu2022petrv2} & 1600 $\times$ 640   & 6                      & 0.421 & 0.524 & 0.681 & 0.267 & 0.357 & 0.377 & 0.186 \\
Sparse4D \cite{lin2022sparse4d} & 1600 $\times$ 640 & 4                 &   {0.436} & 0.541 & 0.633 & 0.279 & 0.363 & 0.317 & 0.177\\ % Sparse4d Table 2
OCBEV \cite{qi2024ocbev}& 1600 $\times$ 900   & 4                      & 0.417 & 0.532 & 0.629 & 0.273 & 0.339 & 0.342 & 0.187 \\
% CAPE-T \cite{xiong2023CAPE} & 1600 $\times$ 640 & 2 & 0.431 & 0.533 & - & - & - & - & - \\
DFA3D-MvACon  \cite{liu2024mvacon} & 1600 $\times$ 900 & 4 & 0.432 & 0.535 & 0.664 & 0.275 & 0.344 & 0.323 & 0.207 \\ % MvACon Table 1

% SOLOFusion & ResNet101 & 1408 $\times$ 512 & 16+1                          & 0.483 & 0.582 & \textbf{0.503} & 0.264 & 0.381 & \textbf{0.246} & 0.207 \\ 
% BEVDet4D-Base (yizhe: seems unnecessary) \cite{huang2022bevdet4d}  & 90${}^\dagger$ & 1600 $\times$ 640  & 2                   & 0.421 &   {0.545} & 0.579 & \textbf{0.258} & \textbf{0.329} & \textbf{0.301} & 0.191 \\ % 

\rowcolor[gray]{.9} 
DuoSpaceNet (Ours) & 1600 $\times$ 640 & 4                                 &\textbf{0.443}  &\textbf{0.547}  &  0.603  &0.275  &0.360  &0.314  &0.195\\ 
\bottomrule
\end{tabular}}
\caption{Comparison on 3D detection results on nuScenes \texttt{val} set. All methods are trained for 24 epochs using ResNet-101-DCN~\cite{zhu2018deformable} and benefit from perspective view pre-training. 
% $\dagger$ indicates equivalent number of training epochs due to data augmentation. 
Test time augmentation is not used for all experiments.
% Results not reported in the original paper are omitted.
}
\label{tab:main_val_set}
\vspace{-0.25cm}
\end{table*}
\section{Experiments}

\noindent \textbf{Dataset.} We benchmark our method on nuScenes dataset~\cite{caesar2020nuscenes}, one of the most widely-used public datasets in autonomous driving. The nuScenes dataset consists of 1,000 driving video clips. Each clip is 20-second long at the sampling frequency of 2Hz. Across the dataset, image data come from the same 6-camera setup, facing at 6 directions, providing a full 360$^{\circ}$ panoramic view. For 3D object detection task, the dataset contains 10 commonly-seen classes (e.g., car, pedestrian), in total $\sim$1.4M bounding boxes. We evaluate our 3D detection results using official nuScenes metrics, including mean average precision (mAP), nuScenes detection score (NDS), mean average error of translation (mATE), scale (mASE), orientation (mAOE), velocity (mAVE) and attribute (mAAE). For map segmentation, we follow previous works~\cite{philion2020lift, xie2022m2bev} and evaluate our method with intersection of union (IoU) metric.

\noindent \textbf{Implementation details.} 
For both single-frame and multi-frame 3D detection experiments, unless specified otherwise, we follow the hyperparameter settings in~\cite{lin2022sparse4d}, including learning rate and schedule, data augmentation, loss functions, and anchor initialization.
 For full model experiments, the BEV feature map is sized $200\times200$, the number of duo space queries is $900$ and the number of decoder layers is 4. All layers have identical settings with $8$ attention heads in both self-attention and cross attention layers. For deformable cross attention layers, we compute $16$ offsets per query. For multi-frame experiments, we use 4 adjacent frames (including the current frame) as temporal input. For all ablation studies, we use ResNet-50~\cite{he2016resnet}, $100\times100$ BEV feature map (if applicable), $800\times320$ input images and a 2-layer decoder, trained for 12 epochs. 
 For map segmentation, we follow PETRv2~\cite{liu2022petrv2} to transform map layers from the global reference frame into the ego frame, and generate two $200 \times 200$ ground truth segmentation masks for $drivable \  area$ and $lane \ boundary$ respectively. 

\subsection{3D Object Detection Results}

\begin{table}
\centering
% \caption{Comparison on model complexity in terms of the number of parameters (params) and the number of floating-point operations (flops). 
% % The input is $1600\times640$ for all models. $\ast$ indicates a lite version of our model for performance-efficiency trade-offs, with the size of BEV feature map reduced to $100\times100$ and without 3D Refinement Module. %0.383 0.455
% }
% \label{tab:main_flops}
\resizebox{0.35\textwidth}{!}{
\setlength{\tabcolsep}{2.0pt}
\begin{tabular}{c|c|c|c|c|c} 
\toprule
\textbf{Method} & \textbf{Space} & \textbf{Params} & \textbf{Flops} & \textbf{mAP}$\uparrow$  &\textbf{NDS}$\uparrow$  \\
\toprule
BEVDet & BEV & 69.5M & 1498.8G & 0.339 & 0.389 \\ 
BEVFormer-S & BEV & 66.6M & 1705.5G & 0.375 & 0.448 \\ 
$\text{Sparse4D}_{T=1}$ & PV & 58.3M & 1453.8G & 0.382 & 0.451 \\ 
% ${}^\ast$
\rowcolor[gray]{.9} 
DuoSpaceNet & BEV+PV & 64.8M & 1771.7G & \textbf{0.399} & \textbf{0.462} \\ 
\bottomrule
\end{tabular}
} 
\caption{Comparison on model complexity in terms of the number of parameters and the number of floating-point operations. 
% The input is $1600\times640$ for all models. $\ast$ indicates a lite version of our model for performance-efficiency trade-offs, with the size of BEV feature map reduced to $100\times100$.
}
\label{tab:main_flops}
\end{table}
 
% \begin{table}[t]
% \centering
% \caption{Map segmentation on the nuScenes \texttt{val} set. 
% % Comparison of temporal input and joint training for detection and segmentation tasks.
% }
% \label{tab:map_seg}
% \resizebox{0.475\textwidth}{!}{
% \setlength{\tabcolsep}{2.0pt}
% \renewcommand{\arraystretch}{1.3}
% \begin{tabular}{c|c|c|c|c} 
% \toprule
% \textbf{Method} & \textbf{Temporal} & \textbf{Joint Training} & \textbf{IoU-Drivable}$\uparrow$ & \textbf{IoU-Lane}$\uparrow$\\
% \toprule
% LSS         & \ding{55} & \ding{55} & 72.9 & 20.0 \\
% \hline
% \multirow{2}{*}{M$^2$BEV} & \ding{55} & \ding{52} & 75.9 & 38.0 \\
%                           & \ding{55} & \ding{55} & 77.2 & 40.5 \\
% \hline
% \multirow{2}{*}{BEVFormer-S} & \ding{55} & \ding{52} & 77.6 & 19.8 \\
%                               & \ding{55} & \ding{55} & 80.7 & 21.3 \\
% \hline
% \multirow{2}{*}{BEVFormer}   & \ding{52} & \ding{52} & 77.5 & 23.9 \\
%                             & \ding{52} & \ding{55} & 80.1 & 25.7 \\
% \hline
% UniAD       & \ding{52} & \ding{52} & 69.1 & 31.3 \\
% \hline
% PETRv2      & \ding{52} & \ding{55} & \textbf{83.3} & 44.8 \\
% \hline
% \rowcolor[gray]{.9} 
% DuoSpaceNet   & \ding{55} & \ding{52} & 80.8 & 45.9 \\
% \rowcolor[gray]{.9}
% (Ours)        & \ding{55} & \ding{55} & 81.2 & \textbf{46.5} \\
% \bottomrule
% \end{tabular}
% }
% \vspace{-0.4cm}
% \end{table}

\begin{table}[t]
\centering
% \caption{Map segmentation on the nuScenes \texttt{val} set. 
% % Comparison of temporal input and joint training for detection and segmentation tasks.
% }
% \label{tab:map_seg}
\resizebox{0.3\textwidth}{!}{
\setlength{\tabcolsep}{2.0pt}
\renewcommand{\arraystretch}{1.3}
\begin{tabular}{c|c|c|c} 
\toprule
\textbf{Method} & \textbf{Joint Training} & \textbf{IoU-Drivable}$\uparrow$ & \textbf{IoU-Lane}$\uparrow$\\
\toprule
LSS          & \ding{55} & 72.9 & 20.0 \\
\hline
\multirow{2}{*}{M$^2$BEV}  & \ding{52} & 75.9 & 38.0 \\
                           & \ding{55} & 77.2 & 40.5 \\
\hline
\multirow{2}{*}{BEVFormer-S}  & \ding{52} & 77.6 & 19.8 \\
                            & \ding{55} & 80.7 & 21.3 \\
\hline
\rowcolor[gray]{.9} 
DuoSpaceNet   & \ding{52} & \textbf{80.8} & \textbf{45.9} \\
\rowcolor[gray]{.9}
(Ours)        & \ding{55} & \textbf{81.2} & \textbf{46.5} \\
\bottomrule
\end{tabular}
}
\caption{Map segmentation on the nuScenes \texttt{val} set. 
% Comparison of temporal input and joint training for detection and segmentation tasks.
}
\label{tab:map_seg}
\vspace{-0.4cm}
\end{table}

Our 3D detection results on nuScenes \texttt{val} set are shown in \cref{tab:main_val_set}. Compared with other state-of-the-art single-/multi-frame methods, our method consistently outperforms others on mAP. Specifically, we achieve 1.7\% mAP gain over the state-of-the-art PV-based method Sparse4D~\cite{lin2022sparse4d} and 2.4\% mAP gain over the state-of-the-art BEV-based method BEVFormer-S~\cite{li2022bevformer}, using the single-frame setup. The same is true for multi-frame results. Among all methods, DuoSpaceNet achieves the lowest mATE by a large margin, suggesting that our duo space design helps the model understand 3D scenes better. When it comes to other metrics, although our method does not achieve 1st place for some entries, we argue that on average our model surpasses others based on the NDS measurement. We also report our results on nuScenes \texttt{test} set in \cref{tab:main_test_set}. Compared with PolarFormer-T~\cite{jiang2023polarformer}, DuoSpaceNet achieves a considerable 1.2\% mAP gain and 2.6\% NDS gain. Note that different methods use different training strategies on the \texttt{test} set (e.g., longer training schedules, more temporal frames, etc.). Nonetheless, our model is capable of achieving competitive results against other state-of-the-art models.

\begin{table*}[!h]
\centering
% \caption{Comparison on 3D detection results on nuScenes \texttt{test} set. All experiments are camera-only methods using pretrained V2-99~\cite{lee2020centermask} backbone. Test time augmentation is not used for all experiments.}
% \label{tab:main_test_set}
\tiny
\resizebox{0.8\textwidth}{!}{
\setlength{\tabcolsep}{3.5pt}
\begin{tabular}{l|c|c|c|c|c|c@{\hspace{1.0\tabcolsep}}c@{\hspace{1.0\tabcolsep}}c@{\hspace{1.0\tabcolsep}}c@{\hspace{1.0\tabcolsep}}c@{\hspace{1.0\tabcolsep}}} 

\toprule
\textbf{Method} & \textbf{Temporal} & \textbf{Image Size} & \textbf{mAP}$\uparrow$  &\textbf{NDS}$\uparrow$  & \textbf{mATE}$\downarrow$ & \textbf{mASE}$\downarrow$   &\textbf{mAOE}$\downarrow$   &\textbf{mAVE}$\downarrow$   &\textbf{mAAE}$\downarrow$ \\
\midrule
DETR3D \cite{wang2022detr3d} & \ding{56} &  1600 $\times$ 900 & 0.412 & 0.479 & 0.641 & 0.255 & 0.394 & 0.845 & 0.133 \\

BEVDet \cite{huang2021bevdet} & \ding{56} & 1600 $\times$ 900 & 0.424 & 0.488 & 0.524 & 0.242 & 0.373 & 0.950 & 0.148 \\ % 
BEVFormer-S \cite{li2022bevformer} & \ding{56} & 1600 $\times$ 900 & 0.435 & 0.495 & 0.589 & 0.254 & 0.402 & 0.842 & 0.131 \\ % 
PETR \cite{liu2022petr} & \ding{56} & 1408 $\times$ 512 & 0.441 & 0.504 & 0.593 & 0.249 & 0.383 & 0.808 & 0.132\\ % PETR Table 1
PolarFormer  \cite{jiang2023polarformer} & \ding{56}   & 1600 $\times$ 900 & 0.455 & 0.503 & 0.592 & 0.258 & 0.389 & 0.870 & 0.132 \\
\rowcolor[gray]{.9} 
DuoSpaceNet (Ours) & \ding{56}  & 1600 $\times$ 640 & \textbf{0.460} &\textbf{0.519}  &0.559  & 0.259  &0.399  &0.765 & 0.134\\ 
\midrule
UVTR \cite{li2022unifying} & \ding{52}   & 1600 $\times$ 900 & 0.472 & 0.551 & 0.577 & 0.253 & 0.391 & 0.508 & 0.123 \\
BEVFormer \cite{li2022bevformer} & \ding{52}   & 1600 $\times$ 900 & 0.481 & 0.569 & 0.582 & 0.256 & 0.375 & 0.378 & 0.126 \\
PETRv2 \cite{liu2022petrv2} & \ding{52}  & 1600 $\times$ 640  & 0.490 & 0.582 & 0.561 & 0.243 & 0.361 & 0.343 & 0.120 \\
PolarFormer-T  \cite{jiang2023polarformer} & \ding{52}   & 1600 $\times$ 900 & 0.493 & 0.572 & 0.556 & 0.256 & 0.364 & 0.439 & 0.127 \\
MV2D \cite{wang2023object} & \ding{52}  & 1600 $\times$ 900 & 0.511 & 0.596 & 0.525 & 0.243 & 0.357 & 0.357 & 0.120 \\
Focal-PETR-T \cite{wang2023focalpetr} & \ding{52} & 1600 $\times$ 640  & 0.511 & 0.592 & 0.546 & 0.243 & 0.373 & 0.357 & 0.115 \\
\rowcolor[gray]{.9} 
DuoSpaceNet (Ours) & \ding{52}  & 1600 $\times$ 640 & \textbf{0.513} & \textbf{0.601}  & 0.508  & 0.254  &0.362  & 0.312 & 0.118 \\ 
\bottomrule
\end{tabular}}
\caption{Comparison on 3D detection results on nuScenes \texttt{test} set. All experiments are camera-only methods using pretrained V2-99~\cite{lee2020centermask} backbone. Test time augmentation is not used for all experiments.}
\label{tab:main_test_set}
\vspace{-0.25cm}
\end{table*}

We also compare our model complexity against other state-of-the-art BEV-only or PV-only methods using input images of size $1600\times640$ under the single-frame setting. 
For all models, we test them on the same machine using DeepSpeed Flops Profiler\footnote{https://www.deepspeed.ai/tutorials/flops-profiler/}. As shown in \cref{tab:main_flops}, under similar model sizes, DuoSpaceNet significantly outperforms BEVDet~\cite{huang2021bevdet} and BEVFormer-S. It is also slightly better than Sparse4D, yet still capable of handling dense segmentation tasks. 

\subsection{Map Segmentation Results}
In \cref{tab:map_seg}, we benchmark the map segmentation performance on nuScenes \texttt{val} set. All methods use ResNet-101-DCN backbone except for M${}^2$BEV, who has a more advanced backbone. Compared with previous methods, our model achieves the highest IoU for both $drivable\ area$ and $lane\ boundary$, regardless of whether the segmentation branch is trained jointly with object detection or not. While sharing modules between different tasks saves computational resources and reduces inference time, the jointly trained model does show a slight performance drop compared to the individually trained model on map segmentation. This phenomenon, known as the negative transfer effect, is consistent with findings from previous studies~\cite{li2022bevformer}.

\subsection{Ablation Studies}

\begin{table}
\centering
% \caption{\zheh{placeholder, real results pending} Ablation of proposed 3D refinement layer. All experiments in this table is conducted with $decoder\_layer=6$.}
% \caption{Ablation of using duo space features.}
% \label{tab:abla_duo_space}
\tiny
\resizebox{0.25\textwidth}{!}{
\setlength{\tabcolsep}{2.0pt}
\begin{tabular}{c|c|c|c|c} 
\toprule
\textbf{Method} & \textbf{w/ BEV} & \textbf{w/ PV} & \textbf{mAP}$\uparrow$  &\textbf{NDS}$\uparrow$  \\
\toprule
BEV Only &\ding{52} & & 0.203 & 0.264  \\
PV Only & & \ding{52} & 0.212 & 0.261 \\
% \midrule
% PV &  &  0.3145 & 0.2333 \\
% Duo (Ours) &\ding{52} &\ding{52} & 0.3318 & 0.2558 \\
% Duo &\ding{52} & 0.3358 & 0.2643  \\
% \rowcolor[gray]{.9} 
Duo (Ours) &\ding{52} & \ding{52} & \textbf{0.216} & \textbf{0.288} \\
\bottomrule
\end{tabular}
}
\caption{Ablation of using duo space features.}
\label{tab:abla_duo_space}
\vspace{-0.2cm}
\end{table}

\paragraph{Duo Space Features.}
To demonstrate the advantages of using BEV and PV features together, we compare the model equipped with our proposed duo space object queries to two baselines where object queries solely attend to either BEV or PV features. As shown in \cref{tab:abla_duo_space}, using features from both spaces leads to a 0.4\% gain in mAP from the PV-only baseline and a considerable 2.4\% gain in NDS from the BEV-only baseline.

\vspace{-0.3cm}
\begin{table}
\centering
% \caption{\zheh{placeholder, real results pending} Ablation of proposed 3D refinement layer. All experiments in this table is conducted with $decoder\_layer=6$.}
% \caption{Ablation of the proposed feature divergence enhancement, dubbed ``FDE'' in the table header.}
% \label{tab:abla_3d_refine}
\tiny
\resizebox{0.25\textwidth}{!}{
\setlength{\tabcolsep}{2.0pt}
\begin{tabular}{c|c|c|c} 
\toprule
\textbf{Method} & \textbf{FDE} & \textbf{mAP}$\uparrow$  &\textbf{NDS}$\uparrow$  \\
\toprule
\multirow{2}{*}{BEV Only} & & 0.203 & \textbf{0.264}  \\
 & \ding{52} & \textbf{0.210} & 0.260  \\
% PV &  &  0.3145 `& 0.2333 \\
\midrule
\multirow{2}{*}{Duo (Ours)} &  & 0.216 & 0.288 \\
% Duo &\ding{52} & 0.3358 & 0.2643  \\
% \rowcolor[gray]{.9} 
 &\ding{52} & \textbf{0.229} & \textbf{0.294} \\
\bottomrule
\end{tabular}
}
\caption{Ablation of the proposed feature divergence enhancement, dubbed ``FDE'' in the table header.}
\label{tab:abla_3d_refine}
\vspace{-0.4cm}
\end{table}
\paragraph{Feature Divergence Enhancement.}
To make BEV features more distinctive from PV features, we propose adding feature divergence enhancement (FDE) during BEV feature generation. As shown in \cref{tab:abla_3d_refine}, while adding it to the BEV-only baseline can improve mAP by 0.7\%, it won't yield any help on NDS. Adding FDE in conjunction with our duo space design, however, will significantly improve the mAP by 1.3\% and NDS by 0.6\%, benefiting from the contrastiveness added between BEV and PV features.

\begin{figure*}[ht]
	\centering  
     \includegraphics[width=0.65\linewidth]{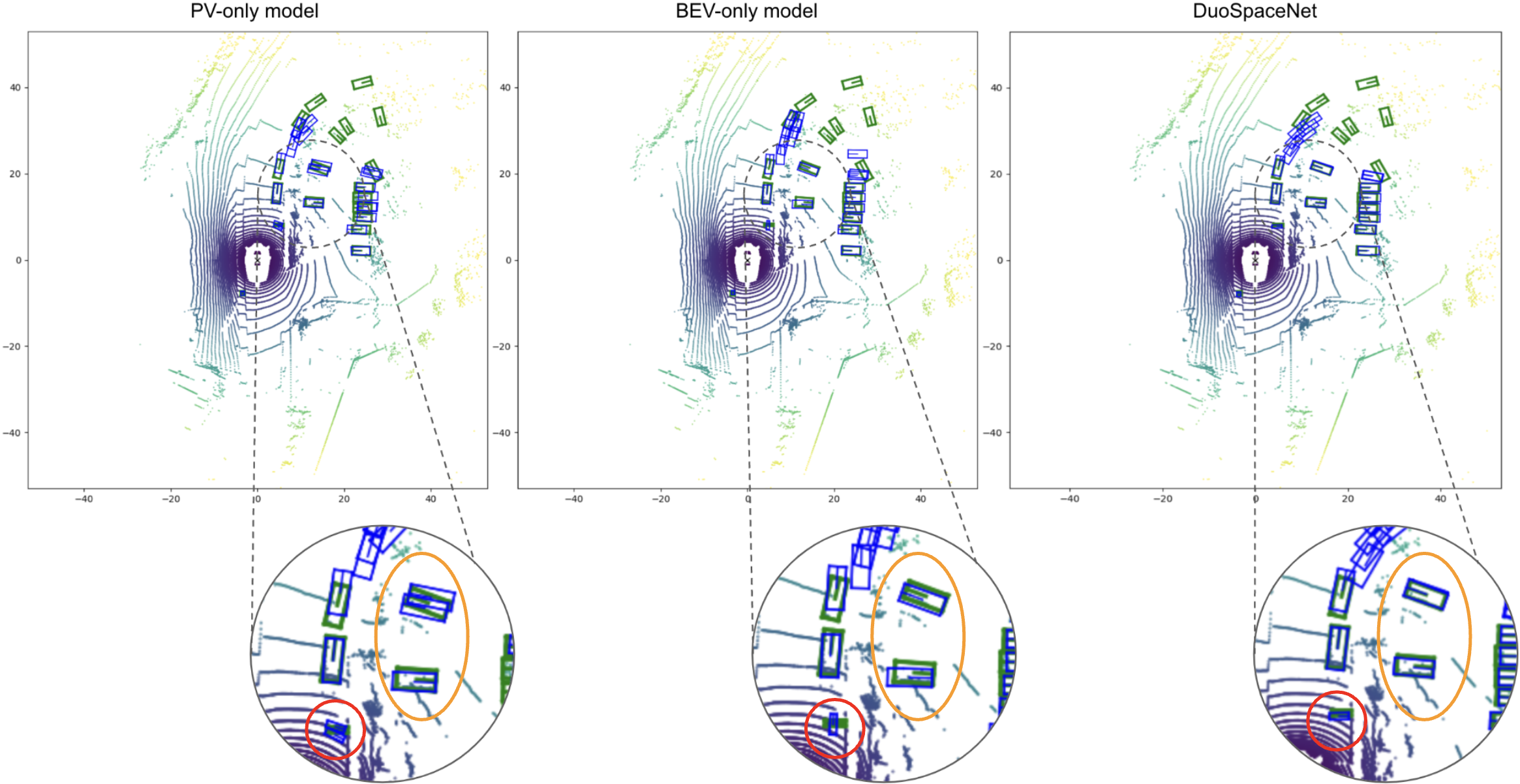}	
 \caption{Qualitative comparison of top‐down 3D detection results among our method, PV‐only, and BEV‐only models. Ground truth bounding boxes are in {\color{ForestGreen} \textbf{green}} and predictions are in {\color{blue} \textbf{blue}}. Our prediction aligns most accurately with the ground truth. Please refer to the supplementary materials for more visualization.}
	\label{fig:vis_bev}
\vspace{-0.4cm} 
\end{figure*}

\vspace{-0.4cm}
\begin{table}
\centering
% \caption{Ablation of the proposed Duo Space Queries.}
% \label{tab:abla_query_design}
\tiny
\resizebox{0.275\textwidth}{!}{
\setlength{\tabcolsep}{2.0pt}

% \begin{tabular}{c|c|c|c|c} 
% \toprule
% \textbf{Method} & \textbf{Shared Pose} & \textbf{Shared Content} & \textbf{mAP}$\uparrow$  &\textbf{NDS}$\uparrow$  \\
% \toprule
% \multirow{3}{*}{Ours} &  &  & & \\
%  &\ding{52} &\ding{52} & 0.225 & 0.290  \\
% % \rowcolor[gray]{.9} 
%  &\ding{52}  & & \textbf{0.229} & \textbf{0.294} \\

% \begin{tabular}{c|c|c|c|c} 
% \toprule
% \textbf{Space} & \textbf{Shared Pose} & \textbf{Shared Content} & \textbf{mAP}$\uparrow$  &\textbf{NDS}$\uparrow$  \\
% \toprule
% \multirow{3}{*}{BEV+PV} &  &  & 0.202 & 0.252 \\
%  &\ding{52} &\ding{52} & 0.225 & 0.290  \\
% % \rowcolor[gray]{.9} 
%  &\ding{52}  & & \textbf{0.229} & \textbf{0.294} \\

\begin{tabular}{c|c|c|c} 
\toprule
\textbf{Shared Pose} & \textbf{Shared Content} & \textbf{mAP}$\uparrow$  &\textbf{NDS}$\uparrow$  \\
\toprule
  &  & 0.202 & 0.252 \\
\ding{52} &\ding{52} & 0.225 & 0.290  \\
% \rowcolor[gray]{.9} 
\ding{52}  & & \textbf{0.229} & \textbf{0.294} \\
 
\bottomrule
\end{tabular}
}
\caption{Ablation of the proposed Duo Space Queries.}
\label{tab:abla_query_design}
\vspace{-0.2cm}
\end{table}
\begin{table}
\centering
% \caption{\zheh{placeholder, real results pending} Ablation of proposed 3D refinement layer. All experiments in this table is conducted with $decoder\_layer=6$.}
% \caption{Ablation of different temporal strategies. }
% \label{tab:abla_temporal}
\tiny
\resizebox{0.3\textwidth}{!}{
\setlength{\tabcolsep}{2.0pt}
\begin{tabular}{c|c|c} 
\toprule
\textbf{BEV Temporal Method} & \textbf{mAP}$\uparrow$  &\textbf{NDS}$\uparrow$  \\
\toprule
\texttt{Recurrent Stacking} & 0.236 & 0.337 \\
\texttt{Learnable Attention} & 0.243 & 0.340 \\
\texttt{Temporal Queries} & \textbf{0.266} & \textbf{0.385} \\
\bottomrule
\end{tabular}
}
\caption{Ablation of different temporal strategies. }
\label{tab:abla_temporal}
\vspace{-0.4cm}
\end{table}
\paragraph{Duo Space Queries.}
Although using feature maps from both spaces inherently has advantages over using those from a single space, optimal performance cannot be achieved without our delicately designed duo space object queries. To validate this, three models differing only in their decoders were obtained. The first model, ``unshared pose and unshared content'', divides classical object queries into two sets, each attending separately to either BEV or PV features in cross-attention layers. The second model, ``shared pose and shared content'', makes each classical object query sequentially pass through self-attention, PV and BEV cross-attention layers, thus sharing pose and content embedding across both spaces. The third model,``shared pose and unshared content'', is equipped with our proposed duo space object queries. As \cref{tab:abla_query_design} reveals, the first setting yields the worst result. In such a setting, each query only sees a single space of features, thereby losing the potential advantages offered by the presence of duo space features.
The performance is marginally improved when pose and content embeddings are both shared, while the best results are achieved coupled with our Duo Space Decoder design. This implies that directly sharing content features across spaces might create latent space confusion due to different feature distributions. Decoupling content embeddings, however, proves to be more optimal by preserving unique feature representations from both spaces.

\vspace{-0.4cm}
\paragraph{Temporal Modeling.}
We demonstrate the necessity of a unified temporal solution for both spaces in contrast to some trivial solutions. We keep using temporal queries in PV space across all experiments. In each experiment, we use a different temporal strategy in BEV space. Specifically, ``\texttt{Recurrent Stacking}'' refers to infusing temporal information by stacking up temporal BEV features. ``\texttt{Learnable Attention}'' refers to infusing temporal information by temporal self-attention proposed in BEVFormer. ``\texttt{Temporal Queries}'' refers to our method where both spaces infuse temporal information into their temporal duo space queries. As clearly shown in \cref{tab:abla_temporal}, temporal strategy matters a lot. The proposed Duo Space Temporal Modeling achieves far superior performance compared with simply using other popular designs.

\vspace{-0.1cm}
\subsection{Qualitative Analysis}
In addition to the quantitative analysis, we report qualitative results of our DuoSpaceNet in comparison with PV-only and BEV-only baselines. In this example, as top-down-view visualization (\cref{fig:vis_bev}) reveals, the bounding boxes (highlighted by the {\color{orange}orange} ellipse) predicted by DuoSpaceNet align more closely with the ground truth in 3D space compared to the PV-only baseline. This alignment is attributed to the more explicit 3D spatial information in BEV features. Regarding the object pose estimation (the {\color{red}red} circle in \cref{fig:vis_bev}, depicting a motorcycle in the real world), due to the significant appearance information in PV features, both DuoSpaceNet and the PV-only baseline accurately predict its heading angle. In contrast, the BEV-only baseline provides an erroneous prediction differing by 90 degrees from the ground truth.
Combined, we are confident that our duo space paradigm achieves ``the best of both world''.

\vspace{-0.1cm}
\section{Conclusion and Limitations}
In this paper, we introduced DuoSpaceNet, a camera-based 3D perception framework that effectively integrates features from both BEV and PV spaces. The proposed duo-space decoder leverages dense BEV representations alongside detailed PV features, significantly enhancing performance on 3D object detection and map segmentation tasks. Extensive experiments on the nuScenes dataset demonstrate that DuoSpaceNet achieves competitive performance compared to state-of-the-art approaches, and ablation studies confirm the effectiveness of each proposed design. Our method can also benefit end-to-end self-driving~\cite{huang2022distributed, villaflor2022addressing}. Beyond autonomous driving, our method holds potential for applications in human-centric tasks, such as human pose estimation \cite{wu2020mebow,wu2022mug} and emotion understanding \cite{wang2023unlocking,wu2023bodily}.

Although our framework theoretically supports long-range 3D detection better than BEV-only methods, full evaluation is limited by insufficient far-range annotations in the nuScenes dataset. Future work will include comprehensive evaluations on larger datasets like Argoverse2~\cite{av2}. Additionally, current map segmentation relies exclusively on BEV features; integrating PV features could further enhance accuracy, particularly for detailed map structures.

\clearpage
{
    \small
    \bibliographystyle{ieeenat_fullname}
    \bibliography{main}
}

\clearpage
\setcounter{page}{1}
\maketitlesupplementary

\begin{figure*}[h]
    \centering  
    \includegraphics[width=0.65\linewidth]{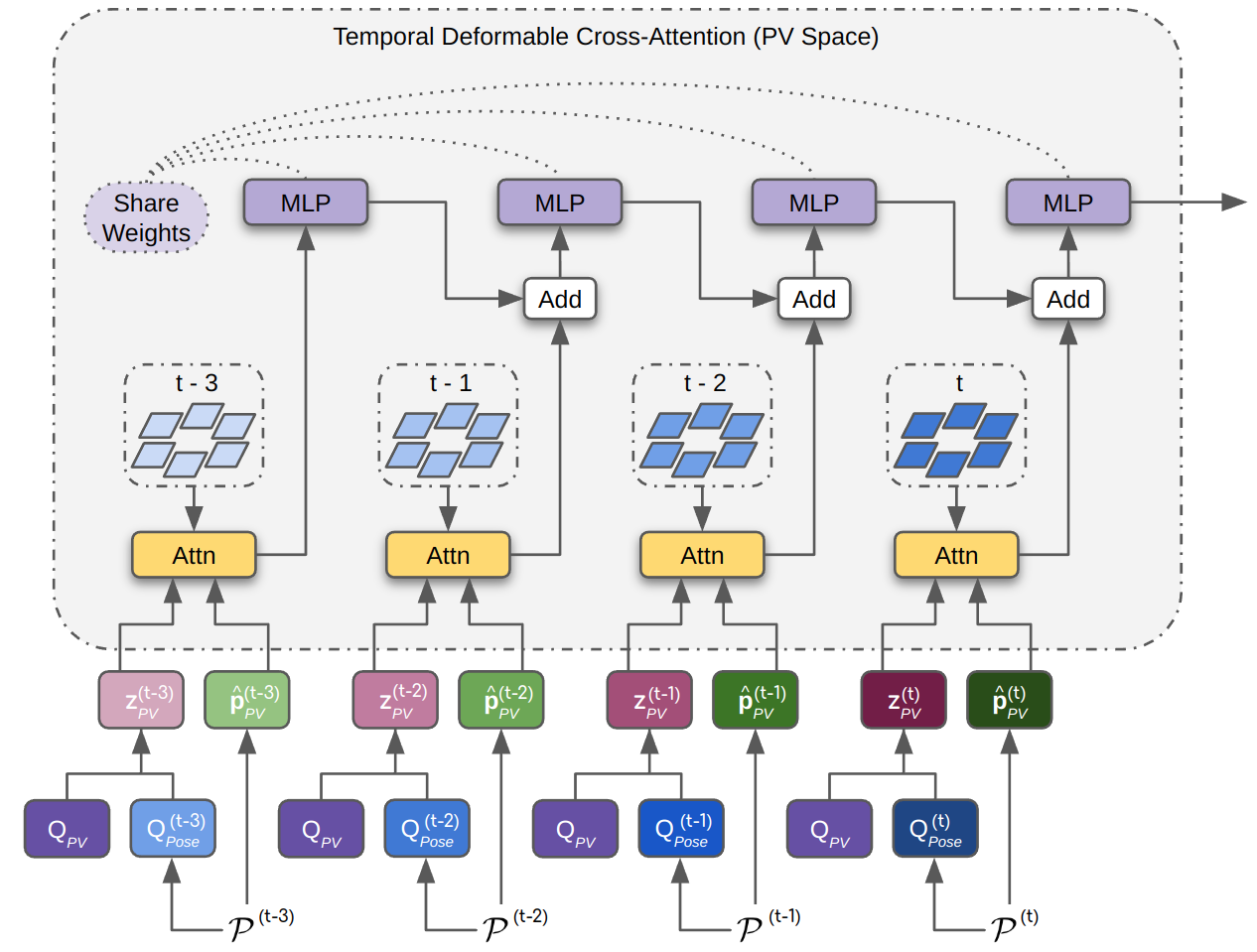}
    \caption{An illustration of space-specific temporal deformable cross-attention in perspective view (PV) space, with $4$ temporal frames. Temporal pose vectors are generated by transforming current the pose vector at current frame to previous frames with motion compensation. After the duo space query composition, duo space temporal queries are formulated for both spaces. Subsequently, in PV space, attention queries, $\mathbf{z}_{PV}^{(\cdot)}$, and their reference points, $\hat{\mathbf{p}}_{PV}^{(\cdot)}$, from timestamp $t-3$ to $t$ are used as input queries for multi-scale deformable attention~\cite{zhu2020deformable}. Each set of queries at a given timestamp only attends to corresponding PV features at the same timestamp. After the attention mechanism, results are aggregated by MLPs in a recurrent manner. Note that weights are shared across all MLPs.
    }
    \label{fig:temporal_attn}
% \vspace{-0.5cm} 
\end{figure*}

\section{Additional details on Duo Space Temporal Modeling.}
Specifically, we generate temporal duo space queries $\mathbf{z}_{BEV}^{(t)}$ and $\mathbf{z}_{PV}^{(t)}$ by infusing past information into shared 3D poses. Assuming the current timestamp is $T$ and the temporal length is $l$ frames, we compute temporal poses $\mathcal{P}_i^{(t)}, i \in \{1, 2, ..., k\}, t \in \{T - l + 1, T - l + 2, ..., T\}$. Ego-motion compensation can be done via a warp transformation matrix from timestamp $t-1$ to $t$, denoted as $[\boldsymbol{R}\,|\,\boldsymbol{t}\,]_{(t)}^{(t-1)}, t \in \{T - l + 1, T - l + 2, ..., T\}$, where $\boldsymbol{R}$ and $\boldsymbol{t}$ refer to the rotational and translational components in the matrix. Object-motion compensation, on the other hand, can be tackled using the predicted velocity of each query, assuming a constant velocity motion model over the $l$-length sequence. Adding up both compensations, we update the object location $x, y, z$ at $t - 1$, dubbed $\mathcal{P}^{(t - 1)}_i\vert_{x, y, z}$, the object orientation $\sin \theta, \cos \theta$ at $t - 1$, dubbed $\mathcal{P}^{(t - 1)}_i\vert_{\theta}$ and the object velocity $vx, vy$ at $t - 1$, dubbed $\mathcal{P}^{(t - 1)}_i\vert_{vx, vy}$, through
\begin{align}
    \mathcal{P}^{(t)}_i\vert_{x', y'} &= \mathcal{P}^{(t)}_i\vert_{x, y} - \Delta t \cdot \mathcal{P}^{(t)}_i\vert_{vx, vy}, \\
    \mathcal{P}^{(t - 1)}_i\vert_{x, y, z} &= [\boldsymbol{R}\,|\,\boldsymbol{t}\,]_{(t)}^{(t-1)}\mathcal{P}^{(t)}_i\vert_{x', y', z}, \\
    \mathcal{P}^{(t - 1)}_i\vert_{\theta} &= \boldsymbol{R}_{(t)}^{(t-1)}\mathcal{P}^{(t)}_i\vert_{\theta}, \\ 
    \mathcal{P}^{(t - 1)}_i\vert_{vx, vy} &= \boldsymbol{R}_{(t)}^{(t-1)}\mathcal{P}^{(t)}_i\vert_{vx, vy}, \\ 
    i \in \{1, 2, ..., k\}&, t \in \{T - l + 1, T - l + 2, ..., T\}, \notag
\end{align}
where $\Delta t$ represents the wall-clock time difference between adjacent frames. 
% For each temporal sample, we keep two feature queues with recent $l$ frames of BEV and PV features. 
We compute the temporal pose embedding for each timestamp $t$ as follows:
\begin{align}
    \left(Q^{i}_{Pose}\right)^{(t)} = \xi\left(\texttt{Enc}(\mathcal{P}_i^{(t)})\right).
\end{align}
We then generate $\mathbf{z}_{BEV}^{(t)}$, $\mathbf{z}_{PV}^{(t)}$, $\hat{\mathbf{p}}_{BEV}^{(t)}$ and $\hat{\mathbf{p}}_{PV}^{(t)}$, $t \in \{T - l + 1, T - l + 2, ..., T\}$ according to Eq.~\ref{Eq:z_bev},~\ref{Eq:z_pv},~\ref{Eq:p_bev} \&~\ref{Eq:p_pv} as temporal inputs. 
Finally, after cross-attention layers (Eq.~\ref{Eq:ca_bev} and~\ref{Eq:ca_pv}), we aggregate temporal outputs via 3-layer MLP before the FFN of each decoder layer. An illustration of temporal cross-attention in PV space is shown in~\cref{fig:temporal_attn}. The temporal cross-attention in BEV space is identical expect for the use of BEV space queries and BEV features as input.

\section{Additional details on experiment settings.}
Following~\cite{lin2022sparse4d}, we initialize the $x,y,z$ coordinates of pose vectors using K-Means centroids on nuScenes \texttt{training} set~\cite{caesar2020nuscenes}. For all experiments, we use AdamW optimizer~\cite{loshchilov2017adamw} and a cosine learning rate scheduler~\cite{loshchilov2016cosine}. The initial learning rate for backbone and other modules are 2e-5 and 2e-4, respectively. No data augmentation is used other than the grid mask used in DETR3D~\cite{wang2022detr3d}. The perception ranges for both the $X$ and $Y$ axes are $[-51.2m, 51.2m]$, which are consistent for both the 3D object detection and map segmentation tasks.

When it comes to the loss functions we use to train the 3D detection, we utilize Focal Loss~\cite{lin2017focal} for bounding box classification and L1 Loss for attribute regression. Duo space queries are assigned to their ground truth via Hungarian Matching introduced in DETR~\cite{carion2020detr}. For segmentation, we use a combination of L1 Loss, Cross Entropy Loss and Dice Loss~\cite{sudre2017dice} for each predicted mask.

% \section{Qualitative analysis.}
% In addition to the quantitative analysis reflected by nuScenes metrics, we report qualitative results of our DuoSpaceNet in comparison with PV-only and BEV-only baselines. In the first example (Figure~\ref{fig:fea2_bev}), its top-down-view visualization reveals that our DuoSpaceNet predicts fewer false negatives than other models at both short and long distances. This suggests that our duo space framework mitigates the uncertainty in depth estimation since it can access visual cues from both BEV and PV spaces.

\begin{figure*}[h]
	\centering  
     \includegraphics[width=0.7\linewidth]{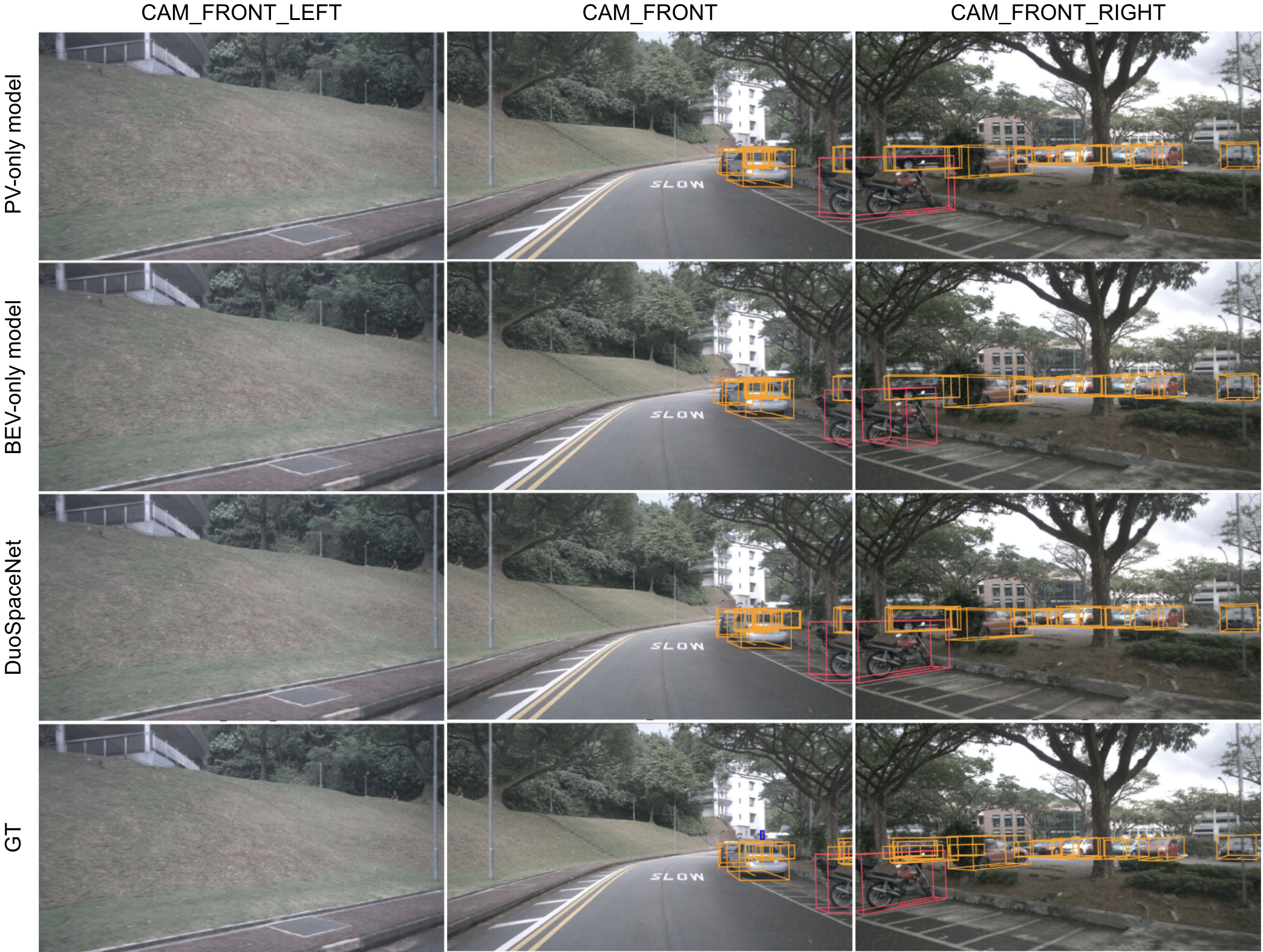}
% \vspace{-0.5cm}	
 \caption{Visualization of 3D detection results in perspective camera view. Different colors represent different categories. Our method achieves the best prediction result for the motorcycle instance w.r.t. its 3D position as well as its orientation, showcasing the effectiveness of incorporating both PV and BEV information in detection queries.}
 % Left camera view was ommited for better visualization.}
	\label{fig:vis_pv}
% \vspace{-0.5cm} 
\end{figure*}

\section{Additional Visualizations}
% In~\cref{fig:vis_pv}, it shows the 3D detection results for the same example as~\cref{fig:vis_bev} in perspective camera view for three different types of detection method, as well as the ground truth. 
In~\cref{fig:vis_pv}, the 3D detection results are displayed in the perspective camera view for the same example as shown in~\cref{fig:vis_bev}, comparing three different detection methods along with the ground truth.

\end{document}